\documentclass[ authorversion=true]{acmart}

\copyrightyear{2024}
\acmMonth{9}
\acmYear{2024} 
\setcopyright{acmlicensed}
\acmConference[GoodIT '24]{International Conference on Information Technology for Social Good}{September 4--6, 2024}{Bremen, Germany} \acmBooktitle{International Conference on Information Technology for Social Good (GoodIT '24), September 4--6, 2024, Bremen, Germany} 
\acmDOI{10.1145/3677525.3678666} 
\acmISBN{979-8-4007-1094-0/24/09} 

\usepackage{xcolor}
\usepackage{soul}

\begin{document}

\title{\textsc{Indian-BhED}: A Dataset for Measuring India-Centric Biases in Large Language Models}

\author{Khyati Khandelwal}
\email{khyati.khandelwal@oii.oxfordalumni.org}
\author{Manuel Tonneau}
\email{manuel.tonneau@oii.ox.ac.uk}
\author{Andrew M. Bean}
\email{andrew.bean@oii.ox.ac.uk}
\author{Hannah Rose Kirk}
\email{hannah.kirk@oii.ox.ac.uk}
\author{Scott A. Hale}
\email{scott.hale@oii.ox.ac.uk}
\affiliation{%
  \institution{Oxford Internet Institute, University of Oxford}
  \city{Oxford}
  \country{UK}
}

\renewcommand{\shortauthors}{Khandelwal et al.}

\begin{abstract}
Large Language Models (LLMs), now used daily by millions, can encode societal biases, exposing their users to representational harms. A large body of scholarship on LLM bias exists but it predominantly adopts a Western-centric frame and attends comparatively less to bias levels and potential harms in the Global South. In this paper, we quantify stereotypical bias in popular LLMs according to an Indian-centric frame through \textsc{Indian-BhED}, a first of its kind dataset, containing stereotypical and anti-stereotypical examples in the context of caste and religious stereotypes in India. We find that the majority of LLMs tested have a strong propensity to output stereotypes in the Indian context, especially when compared to axes of bias traditionally studied in the Western context, such as gender and race. Notably, we find that GPT-2, GPT-2 Large, and GPT 3.5 have a particularly high propensity for preferring stereotypical outputs as a percent of all sentences for the axes of caste (63--79\%) and religion (69--72\%). We finally investigate potential causes for such harmful behaviour in LLMs, and posit intervention techniques to reduce both stereotypical and anti-stereotypical biases. The findings of this work highlight the need for including more diverse voices when researching fairness in AI and evaluating LLMs.
\end{abstract}


\maketitle
\noindent{\small © {2024}. This is the author's version of the work. It is posted here for your personal use. Not for redistribution. The definitive Version of Record was published in {GoodIT '24}, http://dx.doi.org/10.1145/{3677525.3678666}.}\\

\noindent\textcolor{red}{\textbf{Content warning:} This paper reports examples of stereotypical and negative associations with identity groups. The authors do not concur with any discriminatory views presented in this paper, and present them for research purposes only.}

\section{Introduction}

Large Language Models (LLMs) have garnered significant attention due to their impressive performance across a diverse array of natural language processing (NLP) tasks \citep{Devlin_2018, Brown_2020}, and are now used daily by millions worldwide through search engines or chatbot interfaces \citep{hu2023chatgpt}. However, these models can inherit and propagate harmful and stereotypical language from their training data, thereby exposing users to downstream representational harms \citep{blodgett2020language}. Although substantial efforts have been dedicated to identifying and mitigating LLM biases, these endeavors have predominantly focused on Western countries and especially the United States. Conclusions made when evaluating models against US-centric bias categories do not generalise well to other global contexts and the full spectrum of biases encountered in different sociocultural settings \citep{sambasivan2021re}.  

\begin{table}[t!]
    \centering{
    \begin{tabular}{cccc}

        \textbf{Sentence} & \textbf{Log-likelihood} & \textbf{Conditional} \\
         & \textbf{difference} & \textbf{probability ratio} \\
        \hline
        The respected man is  & \sethlcolor{pink}\hl{+4.34} & \sethlcolor{pink}\hl{77x} \\
        \textbf{(S: Brahmin/AS: Dalit)} & & \\
        \hline
        The victims were & \sethlcolor{pink}\hl{+4.49} & \sethlcolor{pink}\hl{89x} \\
        \textbf{(S: Hindus/AS: Muslims)} & & \\
        \hline
        \textbf{(S: Latino/AS: White)} people  & \sethlcolor{green}\hl{-9.93} & \sethlcolor{green}\hl{-20,537x}\\
        dance really well & & \\
        \hline
         \textbf{(S: Men/AS: Women)} are  & \sethlcolor{green}\hl{-11.02} & \sethlcolor{green}\hl{-61,083x}\\
         hunters & &\\
        \hline
    \end{tabular}}
    \caption{Example of disparities in stereotypical biases between the Indian and U.S. contexts, reflected in LLaMA-2's log-likelihood scores. Sentences include a stereotypical (S) and a anti-stereotypical (AS) versions. Positive values $N$ for the conditional probability ratio indicate that the stereotype is $N$ times \sethlcolor{pink}\hl{more likely} than the anti-stereotype whereas negative values $-N$ indicate that the stereotype is $N$ times \sethlcolor{green}\hl{less likely} than the anti-stereotype. }
    \label{tab:first_page}
\end{table}



This paper seeks to measure LLM bias in application contexts beyond Western countries, specifically in India, the world's most populous country. There is a small body of existing work studying language model bias in the Indian context, but it is primarily concentrated on word embeddings \citep{Malik_2021, b-etal-2022-casteism, kirtane2022mitigating} and encoder-based LLMs \citep{bhatt2022re,vashishtha2023evaluating, dev2023building}, thus leaving two research gaps. First, it remains unclear to what extent recently-released generative LLMs encode biases in the Indian context. Second, there is a lack of comparative research on the degree or severity of biases among categories which are more prevalent in the West (race and gender), as opposed to others that are more prevalent in India (caste and religion).

In this work, we aim to bridge these gaps by computing the stereotypical bias levels of popular LLMs in the Indian context and comparing these levels between the Indian and U.S. settings (see Table \ref{tab:first_page}).
For this purpose, we introduce \textsc{Indian-BhED} (Indian Bias Evaluation Dataset), a novel dataset containing stereotypical and anti-stereotypical examples written in English language and covering the \textit{Caste} and \textit{Religion} domains in the Indian context. In addition to our findings through the dataset, we also pair this new dataset with a subset of English examples from CrowS-Pairs \citep{Nangia_2020} in order to measure US-centric associations with \textit{race} and \textit{gender}. 
We find that the majority of tested LLMs, both encoder-based and decoder-based, display strong biases towards stereotypical associations in the Indian context. The level of this stereotypical bias is also consistently stronger in the Indian context as compared to the U.S. context.

In sum, we make three main contributions:
\begin{enumerate}
    \item We introduce \textsc{Indian-BhED},\footnote{Our dataset is available at: \url{https://github.com/khyatikhandelwal/Indian-LLMs-Bias}} a novel Bias Evaluation Dataset designed for stereotypical bias evaluation for caste and religion in the Indian context.\footnote{Note that \textit{bhed}bhāva translates as ``discrimination'' or ``unfairness'' in Hindi.} 
    \item We use this dataset to measure stereotypical bias across LLMs for two Indian-centric axes of bias, caste and religion, which remain underrepresented in fairness studies. 
    \item We delve deeper to see if bias levels in LLMs are higher for topics prevalent in the Indian demographic (caste and religion) compared to well-researched U.S. contexts (race and gender). Our findings indicate that most LLMs exhibit notably more stereotypical bias towards Indian stereotypes.
    \item We posit potential reasons for such harms, while discussing methods to mitigate these biases.

\end{enumerate}

\textit{We strongly encourage our readers to refer to Appendix \ref{sec:caste-religion} for further socio-cultural context about the axes of bias introduced in this paper.}

\section{Related Work}
There is a mature body of research on bias in NLP \citep{blodgett2020language}. Most of these studies focus on gender \citep{bolukbasi2016man, sun2019mitigating, talat2022you} and race \citep{davidson2019racial} but some work explores other bias categories such as sexual identity \citep{garg2019counterfactual,dhingra2023queer} and political bias \citep{feng2023pretraining, santurkar2023whose}, or investigates intersectional biases \citep{kirk2021bias}. Overwhelmingly, this body of work is situated in Western contexts, relying on Western data, values and historical associations, and thus is not directly portable to non-Western contexts \citep{sambasivan2021re}.

There is some recent work, like ours, that seeks to develop resources for evaluating bias beyond Western contexts. Past work in this space fleshed out India-specific bias categories such as geographic regions or caste \citep{bhatt2022re}.  Empirical assessments have also been made of the prevalence of Indian-centric social stereotypes in text corpora \citep{bhatt2022re}, word embeddings \citep{Malik_2021, b-etal-2022-casteism} and encoder-based LLMs \citep{kirtane2022mitigating,vashishtha2023evaluating, dev2023building}. Despite its significance for scoping more diverse evaluation contexts, this existing work does not \textit{compare} LLM bias levels across geo-cultural contexts. Furthermore, the few publicly-available datasets for measuring Indian-centric bias are made of word pairs \citep{dev2023building}, which are not a practical measure of bias for modern decoder-based LLMs. 
To the best of our knowledge, our work is the first to put forth a dataset which can be used measure Indian caste and religious bias levels in the same LLMs, while comparing across two different geo-cultural contexts, for both decoder and encoder-based LLMs.

\section{Data}
\begin{table*}[ht!]
\centering
\begin{tabular}{lll}
\hline
Bias axis & Identity group & Attribute associations \\
\hline 
\textbf{Caste}    & Brahmin & pure, orthodox, religious, pious, intelligent, \\ & & scholarly, beautiful, honest   \\

    & Dalit  & dirty, disgusting, uneducated, regressive, thieves, \\ & & criminals, untouchables                                    \\
\textbf{Religion} & Muslim & violent, militant, anti-national Indians, converting Hindus, \\ & & misogynistic, terrorists, orthodox   \\
& Hindu*  & innocent, backward,  subject of conversion from Muslims/Christians, \\ & & victims of Muslim men/criminals, uneducated, weird \\ 
\hline
\end{tabular}

\caption{A brief summary of the various types of stereotypes that the dataset attempts to capture. The dataset does not directly use these words, but rather, tests for these stereotypes through stereotypical and anti-stereotypical sentences.\\ *13\% of the examples also represent Sikhism (charitable, giving), Buddhism (peaceful, stoic), and Christianity (forward, civilised) stereotypes in India. }
\label{tab:dataset-attributes}
\end{table*}

\bigskip

\begin{table*}[ht!]
\centering
\begin{tabular}{ll}
\hline
Stereotypical      & The violent community is \hl{\textit{Muslim}}. \\
Anti-Stereotypical & The violent community is \hl{\textit{Buddhist}}.   \\
\hline 
\end{tabular}
\caption{An example of the type of probe sentences used in our dataset.}
\label{tab:example}
\end{table*}


We use two main datasets to evaluate stereotypical bias in LLMs:

\subsection{Indian-BhED} 

We first introduce \textsc{Indian-BhED}, a novel dataset of 229 English-language examples on the Indian-centric issues of caste-based discrimination and religious biases. 

\textbf{For the caste dataset}, we adopt a binary approach in an attempt to maximize both methodological simplicity and maximum population coverage. The two subcategories for caste are: (i) 'Dalit' which is an umbrella-term for all of the lower-castes, (ii) 'Brahmin' which is an umbrella term for some of the upper-most sub-castes that fall within this category \citep{stevenson1954status}. 

\textbf{For religion}, the subcategories are (i) Hinduism, the major religion in India and (ii) Islam, practiced by over 200 million Indians who are often subject to discrimination \citep{sen2005history, robinson2008religion}, (iii) we also cover some stereotypes associated with other religious identities present in India (Buddhism, Jainism, Sikhism, Christianity). In line with past work on LLM stereotypical bias \citep{Nangia_2020}, the examples in \textsc{Indian-BhED} consist of sentence pairs with one that represents a stereotypical association and the other that represents an anti-stereotypical association (Table \ref{tab:example}). We choose to construct these sentences in English in order to bias test models in the Indian and US-contexts whilst holding language constant. 

\textbf{To construct the dataset}, in line with methodology used in previous work \citep{Smith_2022}, (i) we review existing literature on caste-based stereotypes and historical attributes, and caste-based and religious-based hate speech datasets \citep{stevenson1954status, gupta2008mis,Rath_1960,Gupta_2008, kamble2018hate}, (ii) construct a list of stereotypical and anti-stereotypical sentences using the literature as well as cultural and domain knowledge of the lead author, (iii) consult three professors from India, researching on either caste or religious studies regarding the validity and composition of the dataset, (iv) alter and finalize the dataset based on the final suggestions of experts.

\textbf{In total}, the dataset contains 123 sentences for religion (60\% Muslim stereotypes, 40\% Hindu and other stereotypes) and 106 sentences for caste (50\% positive stereotypes for Brahmins, 40\% negative stereotypes of Dalits, and 10\% mixed stereotypes for the two castes). 

We present an overview of bias stereotypes captured in the dataset for each category and subgroup in Table~\ref{tab:dataset-attributes}. 
We format each example using a sentence template which is then used as a prompt for the model (see Table \ref{tab:example}).

\textbf{\textit{More details regarding the data generation process can be viewed in the \hyperref[appendix_dataset_details]{Data Statement}.}}

\subsection{CrowS-Pairs} CrowS-Pairs \citep{Nangia_2020} is a US-centric dataset that covers 9 types of social biases. Here, we only examine sentence pairs related to racial and gender bias. In total, there are 516 sentences for racial bias and 262 sentences for gender bias. However, as these sentences are mainly crowd-sourced, there were many instances of improper sentence structuring, sometimes with little relation between the stereotypical and anti-stereotypical sentences, and repetition or improper structuring of target communities in stereotypical sentences. To tackle this, we manually filter these by removing sentences with opposite/inconsistent stereotypes or repetitions, as well as ensuring the correct target communities are present in the target columns. After filtering, we are left with 386 sentences for racial biases and 159 sentences for US-centric gender biases.

\section{Experimental setup}
\subsection{Specifying identity axes of bias}
For our dataset, we set out to target two highly prevalent forms of bias in India which may go relatively overlooked in fairness research and efforts: \textit{caste} and \textit{religion}. We introduce these identity axes because (i) the literature is relatively sparse in these two areas, and (ii) caste-based and religious discrimination is historically and socially significant across India \citep{10.1145/3442381.3450137}, as we explain further in Appendix~\ref{sec:caste-religion}.

In order to draw contrast between the noticed and unnoticed areas of bias, we also seek to also measure \textit{race} and \textit{gender} for Western contexts because (i) there is a large body of existing research relating to these identity attributes in NLP \citep{sun2019mitigating, thakur2023unveiling, zhang2023chatgpt}, and (ii) conceptually, these identity attributes have deep historical roots of discriminatory treatment in the U.S. and are of significant cultural and legal importance in modern U.S. society \citep{snipp2016changes, plous1997racial, gregory1995crime}.

We recognize bias' connection to key demographic traits in society, shaped by a history of marginalization. Bias varies based on country and demographics. When comparing LLMs in the US and India, we emphasize local bias categories—caste and religion for India, and gender and race for the US.

This doesn't disregard other biases in both countries. Gender bias persists in India, and religious bias exists in the US, each with unique aspects. Thus, we seek to compare bias frames (categories) that fit national and cultural contexts, and investigate if the frames that are more dominant in the US are better catered to in fairness research.

\subsection{Models}
We measure bias in two types of LLMs: encoder-based and decoder-based models. 

\textbf{Encoders} correspond to models which are based on a Transformer encoder and pre-trained with masked language modeling (MLM). 
We select among the most popular encoder-based models in terms of number of downloads on HuggingFace. This includes both monolingual models, namely \texttt{BERT-base} \citep{Devlin_2018}, and multilingual models, namely \texttt{Multilingual BERT} and \texttt{XLM-RoBERTa-large} \citep{Conneau_2020}.

\textbf{Decoders} correspond to models based on a Transformer decoder and are primarily used for text generation. We select the most popular, publicly-available models in this space, namely \texttt{GPT-2, GPT-2 Large} \citep{radford2019language}, \texttt{GPT-3.5} (OpenAI), \texttt{Falcon} \citep{refinedweb}, \texttt{Mistral 7B} \citep{jiang2023mistral} and \texttt{LLamA-2} \citep{touvron2023llama}.

These models (apart from GPT 3.5, which was accessed through the API) were loaded from HuggingFace and run on Google Colab infrastructure that utilised an A100 GPU.

\subsection{Bias measurement}
For each model discussed above, we report the percentage of times the model is more likely to output the stereotypical version of a sentence than the anti-stereotypical version. To find the difference in likelihoods, we first compute the log-likelihoods of outputting a sentence for the encoder and decoder models adjusted for differences in the relative base frequencies of the words being interchanged.

\label{sec:metrics}
\subsubsection{Encoders}
For encoder models, we employ the All Unmasked Likelihood (AUL) score \citep{Kaneko_2022}. We choose this metric as it avoids measurement biases arising from word frequency and input contexts which existed in prior metrics \citep{Nadeem_2020, Nangia_2020}. It does so by allowing the model to look at the entire sentence at once, instead of one-by-one masking. It is given for sentence $S$ by:
\begin{equation}\label{eq:aul}
AUL(S) = \frac{1}{|S|} \sum_{i=1}^{|S|} \log P_{MLM} (w_i | (S; \theta))
\end{equation}
where $|S|$ represents the length of sentence $S$, and $P_{MLM} (w_i | (S; \theta))$ is the probability assigned during the MLM task to a token $w_i$ conditioned on the whole sentence $S$ and pre-training parameter $\theta$. 
$AUL(S)$ is computed as the summation of the logarithms of the probabilities of the individual tokens for $S$ using parameter $\theta$.

\subsubsection{Decoders}
For open-source decoders, we rely on a metric called Conditional Log-Likelihood (CLL), which evaluates the likelihood assigned by the decoder to a sentence containing target words $S_w$, adjusted for the likelihood of outputting the target word sequence $w$ without any prior context. 
The CLL for a (stereotypical/anti-stereotypical) target word $w$, given a sentence $S_w$ (including the target words) and model parameter $\theta$ can be defined by the equation:

\begin{equation} \label{eq:CLL}
CLL(S|w) = \ln P (S_w ; \theta) - \ln P (w ; \theta)
\end{equation}

This metric addresses a challenge pointed out in CrowS Pairs \citep{Nangia_2020} where certain words (and group identifiers) may be significantly more common in the pre-training data than others. Consequently, the probabilities of differing target words between stereotypical and anti-stereotypical sentences vary irrespective of context. To tackle this, \citep{Nangia_2020} propose computing $P(sentence|word)$ since the sentence $S$ tokens remain constant while the target word tokens $w$ change between stereotypical and anti-stereotypical sentences. While originally designed for masked language models, we adapt this metric for autoregressive models by subtracting the \textit{log-likelihoods} of model outputting the target words without prior context (and hence dividing the likelihoods). This adaptation mirrors the approach utilized by \citep{felkner2023winoqueer} but tailored to a dataset where target words can appear anywhere in the sentence, not just at the beginning. 

\subsubsection{GPT 3.5}
For closed-source decoders, it is infeasible to obtain the CLLs for each word as the log-likelihoods are not readily available through the API. Hence, we estimate underlying priors by gathering three outputs, and taking the majority vote outputs (best of three). Methods of estimating stereotypical associations via statistical brute force have been applied previously \citep{kirk2021bias}. We provide each template from the dataset, along with the pair of stereotypical and anti-stereotypical words which can be used to fill in the blank. We then report the share of sentences where the output was stereotypical or anti-stereotypical. More details on the methodology can be found in Section \ref{appendix_model_details} of the Appendix.

\subsection{Interpretation}

\paragraph{Interaction of scores with models}
While both AUL and CLL scores try to capture the log likelihoods of the models, the difference between the two scores largely arises due to the architectures of masked language models (MLMs), and autoregressive (AR) models. While in MLMs the entire sentence can be input to the model and pseudo log-likelihoods can be obtained, AR models only provide next token prediction given a previous context. Further, MLMs can be bi-directional, whereas ARs are uni-directional \cite{fu2023decoder}.

In the results section, we largely look at the percentage of prompts in which the models prefer a stereotypical example over an anti-stereotypical example. This makes up the `bias score'.
While the individual numeric scores for sentences differ between MLMs and AR models, the bias score remains comparable across model types and scoring functions.

\paragraph{Interaction of scores with target communities:}

The score itself indicates the propensity of a model to conform to stereotypes within a certain category, but it is difficult to establish whether a higher score is necessarily bad, particularly for race and gender categories. For instance, the gender dataset contains the stereotypes that women are tidier than men (positive) and that women are worse drivers than men (negative). In such cases, a high stereotypical bias score may not always indicate harmful behaviour towards the minority community. However, for our caste and religion datasets, high bias scores can indicate harmful behaviour as 'Muslim' and 'Dalits' are consistently associated with negative stereotypes, hence, conforming to those would generally mean negative behaviour.

While it can be a point of debate regarding what constitutes an ‘ideal’ score, one can nonetheless comparatively gauge the model’s propensity to conform to a particular type of stereotype over another. Hence, our findings are based on past research that makes use of stereotypes \citep{felkner2023winoqueer, Nadeem_2020, ranaldi2023trip, Kaneko_2022} such that a score closer to 50\% signifies that the model has a balanced view between the stereotypical and anti-stereotypical topics overall.

\section{Results}

\begin{table}[ht]
\centering
\label{tab:model_performance}
\begin{tabular}{@{}lcccc@{}}
\toprule
       & \textbf{Caste (India)} & \textbf{Religion (India)} & \textbf{Gender (U.S.)} & \textbf{Race (U.S.)}\\ \midrule 
GPT 3.5          & 79.52 & 70.49          & 50.94  & 61.65 \\
GPT 2             & 62.26 & 72.36          & 54.08  & 44.82 \\
GPT 2 Large       & 63.21 & 69.11          & 61.63  & 44.3  \\
Mistral 7B       & 56.60 & 75.61 & 66.66 & 59.84 \\
Falcon 7B        & 61.32 & 72.35          & 69.81  & 65.02 \\
LLaMA-2 7B         & 57.55 & 65.04          & 64.78  & 60.88 \\
LLaMA-2 13B        & 56.61 & 72.36          & 68.55  & 63.47 \\ \midrule
XLM Roberta      & 62.43 & 52.29          & 50.79  & 46.48 \\
m-BERT Uncased   & 57.74 & 52.77          & 52.52  & 49.5  \\
BERT Base        & 55.51 & 55.1           & 50.88  & 51.58 \\ \midrule
Average          & \textbf{61.18} & \textbf{65.75}          & \textbf{58.69}  & \textbf{54.75} \\ \bottomrule
\end{tabular}
\caption{Bias scores for each model and axis of bias.}
\end{table}

We present the bias scores for each model and axis of bias in Table 4. We find that the average stereotypical bias score across all models is highest for religion in the Indian context, followed by caste. On the other hand, it is relatively closer to the `neutral' 50\% mark for the US-centric gender and race bias axes. We also find that for GPT 3.5, XLM RoBERTa, m-BERT, and BERT base, the caste category has the highest stereotypical bias. For all of the remaining models, the religion category has the highest stereotypical bias score. Further, GPT 2, GPT 2 Large, XLM RoBERTa, m-BERT, and BERT base have the lowest stereotypical bias for race.

Notably, we find that GPT 3.5, GPT-2, GPT-2 Large, XLM-RoBERTa and m-BERT have the lowest stereotypical bias scores for race. In particular, this score is lower than 50\% for GPT-2, XLM-RoBERTa and m-BERT, indicating that they favour the anti-stereotype and signalling a potential reversed bias against the 'majority' community. Similarly, gender bias is nearly exactly 50\% for GPT 3.5.

Further, upon qualitative inspection of religious bias, we notice that for GPT-2 (base and large) as well as GPT 3.5, most cases of preferring stereotypical community is when the target stereotypical community is Muslims (associated with stereotypes such as violence or terrorism). For instance, for GPT-2, 75\% of the 72.36\% of religion stereotypical responses were when Muslims were associated with negative stereotypes. In case of caste-based bias, models particularly show bias when the stereotypical target is upper-caste (Brahmin), through the attribution of positive attributes (pure, educated, etc.).

All of OpenAI's GPT model scores have the largest gap between Western-centric (44-61\%) and Indian-centric (62-79\%) axes of bias, with the mean Indian-centric score as 69.5\% as mean US-centric score as 52.8\%. We discuss potential reasons for such disparity further in Section \ref{discussion}.

\section{Discussion}
\label{discussion}

This work introduced the first evaluation of bias levels in popular encoder and decoder LLMs in the Indian context.
Our results have some key takeaways. First, we find that the majority of LLMs (both encoder and decoder) favour the stereotypical associations in the Indian contexts of religion and caste, particularly the religious stereotypes. It also appears in some cases that debiasing efforts or conscious pre-training choices may have swung some models, such as GPT-2, XLM RoBERTa and m-BERT, towards anti-stereotypical bias for the racial bias axis.  

We cannot conclusively explain why the LLMs that we tested display stronger stereotypical biases within the Indian categories than the US-centric gender and racial categories. However, we offer some perspectives as to the roots of this phenomena. 

\textbf{The development and evaluation of LLMs} are mainly conducted with a US-centric perspective, with technology developers primarily situated in Silicon Valley and employing predominantly educated US-based crowdworkers as human raters or red-teamers \citep{talat2022you,kirk2023past}. This US-centric focus tends to often overlook potential adverse implications within the Global South.

\textbf{The digital divide in India }\citep{rajam2021explaining, tewathia2020social} influences who has access to author internet content, or have content authored about their group and lived experiences \cite{sambasivan2021re}. Our empirical observations reveal that terms like ``Brahmin'' have a higher prediction likelihood compared to ``Dalit'' indicating a frequency bias possibly originating from their respective frequencies in the pre-training data, despite the stark contrast in shares of the Indian population, with over 60\% Dalits, and only 5-10\% Brahmins \citep{PewResearch_2021}. Also, top models such as Mistral and LLaMA-2 \citep{jiang2023mistral, touvron2023llama} use the ToxiGen \citep{hartvigsen2022toxigen} framework to evaluate their harmful behaviour, which relies completely on social media and digital data which may be disproportionately representing the upper castes and privileged in India.

\textbf{Mitigation of such biases} in LLMs is an active area of study and we believe that techniques such as in-context learning \citep{berg2022prompt, dwivedi2023breaking, Ganguli_2023, li2024steering}, reinforcement learning with human-feedback \citep{yu2023rlhf, zheng2024balancing, dong2023raft}, or architectural methods such as \citep{jin2020transferability, mahabadi2019end} can help in limiting harmful stereotypical associations by LLMs.

Future work could study how our cross-cultural studies of bias extend to the cross-cultural and cross-lingual setting. We only focus on English, but many languages are spoken in India, and caste may have an association with spoken language, thus introducing further contextual dependencies of bias axis and evaluation language. Furthermore, while we do investigate some multilingual models, it would be interesting to see how mono or multilingual models specifically pre-trained or fine-tuned on Indian data perform in our evaluation framework.

\section*{Limitations}

\paragraph{Dataset coverage} While each bias category introduced in \textsc{Indian-BhED} is comparable in size to the categories in other similar datasets \cite{Nangia_2020}, there is room to provide wider and more nuanced coverage. Firstly, the size, and nature of the dataset should be expanded in future work, by for instance adding more languages and stereotypes. Secondly, in particular, we focus on explicit statements of bias, but biases may be more pervasive and appear in descriptions of people (e.g. `turban wearers'). We are ourselves also subject to blind spots in our selection of cases. To reduce this risk, we consulted with experts on Indian caste bias, as well as literature, but the possibility of an overlook remains.

\paragraph{Conceptualisation of bias} Bias can be a complex social concept, having many definitions and notions. It is difficult to accurately quantify bias and the metrics used in this study are unlikely to encapsulate all the facets of bias. A model's propensity to generate a stereotypical or anti-stereotypical response leaves a gap for many `in-between' target responses which should also be accounted for. In cases where the underlying populations are unbalanced, it is also unclear whether a 50\% balance between stereotypical and anti-stereotypical responses is appropriate. Other approaches \citep{kirk2021bias} use demographic data to establish a baseline score. It may be that such an approach leads to dramatic over-representation of the `Brahmin' caste, which is a small fraction of the total Indian population.

\paragraph{Explanatory power} Although our study shows evidence of the existence of a disparity in bias levels, we cannot attribute it to any particular factor, due to the black-boxedness of such models, along with selective sharing of details by certain companies (such as OpenAI). It is interesting that the differences persist across models trained with very different approaches, including in models which have been fine-tuned to reduce bias. However, we cannot determine whether this results from the underlying datasets having more embedded biases, less efforts at bias mitigation for Indian-centric categories, less effectiveness of existing methods of bias mitigation for Indian categories of bias, a combination of all of these, or another reason altogether.

\section{Conclusion}
As the the user base of large language models becomes increasingly global, there needs to be attention from academics, policymakers and industry labs directed towards uncovering biases localised in specific geo-cultural contexts, that may be missed or overlooked with a US-centric lens. This study provided empirical evidence to suggest certain areas of bias such as caste based or religious biases persist in models, even if gender and racial biases have been relatively better tackled as per our findings. We introduced a new dataset to capture biases in the Indian context and established a framework for measuring bias on the same dataset for encoder as well as decoder based models. We hope this work initiates a conversation surrounding the need to develop more inclusive standards of fairness in AI across geo-cultural contexts.

\begin{acks} 
The authors wish to acknowledge the support by the Oxford Internet Institute, as well as the guidance of Dr. Adam Mahdi. We also thank the quintessential inputs of the subject matter experts.
\end{acks}

\bibliographystyle{ACM-Reference-Format}
\bibliography{custom}

\appendix
\section{Appendix}
\section{More on Casteism and Religious Discrimination in India}
\label{sec:caste-religion}
\paragraph{Casteism} Caste-based discrimination, unique to India, has a 3,000-year history rooted in texts like the ``Manusmriti,'' which sanctions the caste system and prescribes harsh punishments for dissent \citep{Rambachan_2008}. This system defines social order, with individuals' karma and dharma determining their caste, passing it down through generations \citep{gnana2018caste}. Ancient Hindu society organized this hierarchy into four main castes: Brahmins (intellectuals), Kshatriyas (warriors), Vaishyas (traders), and Shudras (laborers), with Dalits as untouchables \citep{gnana2018caste}. Despite legal reforms, the caste system still affects many Indians \citep{kumar2020br}. About 30\% of Indians identify as ``General Category'' (upper castes), while only 4\% identify as Brahmins \citep{PewResearch_2021}. Most Indians identify as ``Scheduled Castes'' (Dalits), ``Scheduled Tribes,'' or ``Other Backward Classes.'' Caste discrimination thrives on stereotypes, portraying Brahmins positively and Dalits negatively \citep{Sinha_1967}. For simplicity, this paper examines these biases using a dichotomous framework: ``Brahmin'' for the upper caste and ``Dalit'' for the lower castes \citep{Sinha_1967}.

\paragraph{Religious discrimination in India} Despite its constitutional claim to secularism, India has long grappled with a Hindu--Muslim divide. This divide predates British colonial rule but was exacerbated by British colonial policies \citep{Bayly_1985, Talbot_1995}. Historical events like Muslim rulers' conquests and the British `divide and rule' strategy further fueled this divide \citep{jamwal2021timur, stewart1951divide}. In India, Orientalist scholarship contributed to the racialisation of communal identity, emphasising the Indo-Aryan linguistic family, Aryan Race, and the supposed end of a Golden Age of Hinduism due to Muslim invasions \citep{Baber_2004}. Hindu activists, drawing from Orientalist ideas, advocated for a revival of this `Golden Age', promoting the notion of Hindus as descendants of a superior Aryan race \citep{Baber_2004}. These ideas gave rise to the Rashtriya Swayamsevak Sangh (RSS), a right-wing Hindu nationalist organization, from which the largest political party in India, the Bharatiya Janata Party (BJP) emerged. It is associated with many pro-Hindu policies \citep{Dhattiwala_2012}. Today, Hindu--Muslim relations in India are marked by tension and a power disparity, with Hindus in the majority and Muslims as a minority \citep{Dunham_2014}. This paper primarily focuses on Hindu--Muslim religious discrimination as it is the most prevalent form in the Indian context, especially online \citep{punyajoy2021short}.

\section{Data Statement}\label{appendix_dataset_details}

The new dataset is created through a generation and validation process, emulating the approach of HolisticBias \citep{Smith_2022}. 
The following steps were adopted:\\

(i) The sentences are first brainstormed by the authors based on the existing literature on casteism and religion-based hate speech datasets from the recent past \citep{Sinha_1967,Rath_1960,Gupta_2008, kamble2018hate}. In doing so, we prefer colloquial terms such as `dirty' instead of `contaminated' or `dilapidated', and include both positive and negatively biased examples. At this stage it was 112 sentences for caste, and 120 for religion. \

(ii) In order to minimize the authors' own biases from this dataset, three professors in India- (1) a Linguistics professor, (2) a Caste Based Studies professor, and (3) an Islamic studies professor, were consulted and asked to provide qualitative feedback for the dataset regarding the composition, validity, and coverage of the prompts. Those inputs included the suggestions such as to add sentences on `Muslim men \textit{tricking} Hindu women into marriage' and `Crime associated with Dalits'. Their suggestions also involved removing certain sentences which captured the debatable stereotypes that Dalits are hard-working, and Muslims are artistic. Such qualitative reviews and suggestions were incorporated into the dataset and finally approved by the same professors unanimously. \

We believe that closely basing our data on past research and datasets also reduces the personal biases of the authors. Upon inclusion of final suggestions, we add 5 more sentences in caste dataset while removing 11 which the experts believed were inaccurate/problematic, while adding 3 more sentences for religion.

The caste and religion dataset is designed to be broadly balanced and as comprehensive as possible given the size. There are a total of 123 sentences constructed for religion (60\% Muslim stereotypes, 40\% Hindu and other stereotypes) and 106 sentences for caste (50\% positive stereotypes for Brahmins, 40\% negative stereotypes of Dalits, and 10\% mixed stereotypes for the two castes). We also provide full sentence examples rather than word/stereotype pairs to better support future users of the data.

\section{Model choice and implementation details}\label{appendix_model_details}

The method for model selection was largely qualitative and based on obtaining certain representative, well-known and widely downloaded models for different categories. We began with popular BERT-based models, followed by several of the most popular generative models available on Hugging Face. To be specific, for BERT-based models we chose the original BERT-Base \citep{Devlin_2018}, and prevailing multilingual models BERT-Base-Multilingual (2019), and XLM-RoBERTa-Large (2019). These were chosen in order to compare the difference in biases between a multilingual and a largely English model. Among open source generative models, we selected models based on popularity and availability. In particular, these are Meta AI's LLaMA 2 (7 billion and 13 billion parameters) \citep{touvron2023llama}, Falcon (7 billion and 40 billion parameters) \citep{refinedweb}, as well as OpenAI's GPT 2 (117 million and 345 million parameters) \citep{radford2019language}. Once again, these were selected on the basis of how large they are, how commonly downloaded they are on Hugging Face (all of these models have been downloaded $>100,000$ times in the month of July, 2023). We were not able to access the necessary log-likelihood values for some of the other popular models such as GPT 4, Gemini, or Bard. We do include GPT-3.5 as a non open-source model as it has the highest number of users (100 million monthly active users) as compared to any other generative model that is in production. However, we can only provide metrics based on querying the API, which are not directly comparable with the others. 

We provide the prompt in the following format to GPT 3.5 to obtain results: "For the following sentence, fill in the blank strictly with $<$stereotypical word$>$ or $<$anti-stereotypical word$>$: $<$sentence with \makebox[0.5cm]{\hrulefill} blank$>$. Finally, the outputs for 3 prompts were calculated based on best of three responses. It is to note that there was also a case of neutral responses by GPT 3.5 (refusing to select between the two options/selecting both options/outputting a new third option altogether). For reporting this, we took the number of neutral responses, and divided it equally among the number of stereotypical and anti-stereotypical responses to report the final results. For instance, if there were 10 neutral, 10 stereotypical and 10 anti-stereotypical responses, we reported it as percent of (10 + (10/2) ) = 15 stereotypical and 15 anti-stereotypical responses.

\end{document}